\pdfoutput=1 
\documentclass{article}

     \usepackage[final, nonatbib]{neurips_bdl2019}

\usepackage[utf8]{inputenc} 
\usepackage[T1]{fontenc}    
\usepackage{hyperref}       
\usepackage{url}            
\usepackage{booktabs}       
\usepackage{amsfonts}       
\usepackage{nicefrac}       
\usepackage{microtype}      
\usepackage[utf8]{inputenc}
\usepackage{amsmath,amsthm,amssymb,amsfonts, color}

\usepackage{hyperref}
\usepackage{lipsum,graphicx,multicol}

\usepackage[utf8]{inputenc}
\usepackage{amsmath,amsthm,amssymb,amsfonts, color}
\usepackage{graphicx,subfig, import}
\usepackage{algorithm,algorithmic}
\usepackage{booktabs,multirow}
\usepackage{fullpage}
\usepackage{pdfpages}
\usepackage{blindtext}
\usepackage{wrapfig}
\usepackage{tabularx} 

\title{Differential Bayesian Neural Nets}

\author{%
	Andreas Look\quad and\quad  Melih Kandemir \\
	Bosch Center for Artificial Intelligence \\
	\texttt{\{andreas.look, melih.kandemir\}@de.bosch.com} \\
}

\begin{document}

\maketitle

\begin{abstract}
	Neural Ordinary Differential Equations (N-ODEs) are a powerful building block for learning systems, which extend residual networks to a continuous-time dynamical system. We propose a Bayesian version of N-ODEs that enables well-calibrated quantification of prediction uncertainty, while maintaining the expressive power of their deterministic counterpart. We assign Bayesian Neural Nets (BNNs) to both the drift and the diffusion terms of a Stochastic Differential Equation (SDE) that models the flow of the activation map in time. We infer the posterior on the BNN weights using a straightforward adaptation of Stochastic Gradient Langevin Dynamics (SGLD). We illustrate significantly improved stability on two synthetic time series prediction tasks and report better model fit on UCI regression benchmarks with our method when compared to its non-Bayesian counterpart. 
\end{abstract}
\section{Introduction}
Deep neural nets are in widespread use of machine learning applications. They owe their unprecedented expressive power to repetitive application of a function that non-linearly transforms the input pattern. Furthermore, if the transformation is designed to be a ResNet module \cite{resnet}, the processing pipeline can be viewed as an ODE system discretized across even time intervals \cite{neural_ode}. Rephrasing this model in terms of a continuous-time ODE is referred to as a Neural ODE. While the generalization capabilities of Neural ODEs have been closely investigated by \cite{neural_ode}, their success as a Bayesian inference building block remains unexplored. In order to answer this question, we devise a generic Bayesian neural model that solves a SDE \cite{oksendal} as an intermediate step to model the flow of the activation maps. Our method differs from earlier work in that we model the drift and diffusion functions of an SDE as Bayesian Neural Nets (BNN), instead of the mean and covariance functions of the Gaussian Process (GP) posterior predictive \cite{diffgp} or a vanilla neural net with a fixed dropout rate global to the synaptic connections \cite{nsde}. Other attempts of coupling SDEs with neural networks consist of finding unknown parameters to otherwise known functions \cite{sde_blackbox}.

The contributions of our work are as follows: i) we build a Neural SDE by assigning two seperate and potentially overlapping BNNs on the drift and diffusion terms, ii) we show how SGLD can naturally be used to infer the consequent model as an alternative to variational inference, iii) and we illustrate how crucial uncertainty-aware learning is for time series modeling with Neural ODEs.

\section{Stochastic Differential Equations}
An SDE can be expressed in the following generic form
\begin{equation}
dx(t) = \mu (x(t),t)dt+\sigma(x(t), t)dW(t).
\end{equation}
The equation is governed by the drift $\mu(x(t))$, which models the deterministic dynamics, and the diffusion $\sigma(x(t))$, which models the stochasticity in the system. Further, $dt$ represents the time increment and $W(t)$ is a Wiener process. There does not exist any closed-form solution to generic SDEs, hence numerical approximation techniques are employed, possibly the most popular of which is the Euler-Maruyama discretization method, which suggests the following update rule
\begin{equation}
x_{i+1} = x_{i} +  \mu (x_i, t_i)\Delta t+\sigma(x_i, t_i)\Delta W ,
\label{eq:euler_maruyama}
\end{equation}
where $\Delta W \sim \mathcal{N}(0, \Delta t)$. The same approximation holds when the variable $x_i$ is a vector $x \in \mathbb{R}^{D}$. In this case the diffusion term is a matrix-valued function of the input and time $\sigma (x_i, t_i) \in  \mathbb{R}^{D \times P}$ and corresponding $\Delta W$ is modeled as $P$ independent Wiener processes $\Delta W \sim \mathcal{N}(0, \Delta t I_P)$, where $I_P$ is a $P$-dimensional identity matrix \cite{oksendal}.

\section{Differential Bayesian Neural Nets}

Assume for brevity that we are given a supervised learning problem, i.e. we aim to find a mapping from inputs $x$ to outputs $y$. We pose the below probabilistic model 
\begin{align*}
\theta_1, \theta_2 &\sim p(\theta_1) p(\theta_2),~~~~~~~~~~~~~~~~~~~~~~~~~~~~~~~~~~~~~~~~~~~~~~~~~~~~~~~~~~~\mathrm{prior}\\
h(t) &\sim  p(h(t)|\theta_1,\theta_2),           ~~~~~~~~~~~~~~~~~~~~~~~~~~~~~~~~~~~~~~~~~~~~~~~~~~~~~~\mathrm{DBNN}\\
y|h(T),x &\sim p(y|h(T)).~~~~~~~~~\mathrm{s.t.}~~~~~h(0) \sim \delta_{x}.~~~~~~~~~~~~~~~~~~~~~\mathrm{likelihood}
\end{align*}

\begin{wrapfigure}{R}{3.5cm}
	\centering
	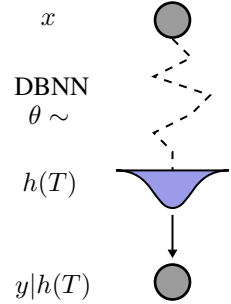
	\caption{Illustration of our algorithm. First an input $x$ is passed through the DBNN. The resulting distribution is then used to calculate $y|h(T)$.\label{fig:example}}
\end{wrapfigure}

The last step above is a likelihood suitable to the learning setup Dirac delta $\delta_{x}$ evaluated on the input observation $x$ and for some chosen $T$ that represent the duration of the flow, namely the model capacity. The critical intermediate step of the model is the stochastic process on the continuous-time activation maps $h(t)$, which we refer to as the {\it Differential Bayesian Neural Net (DBNN)}:
\begin{align*}
p(h(t)|\theta_1,\theta_2) = \int m_{\theta_1}(h(t),t) dt + \int L_{\theta_2}(h(t), t) dB(t)
\end{align*}
, where ${\theta_1}$ and $\theta_2$ are the synaptic weights of BNNs $m_{\theta_1}(\cdot) \in \mathbb{R}^{D}$ on the drift vector and $L_{\theta_2}(\cdot) \in \mathbb{R}^{D \times P}$ on the diffusion matrix, respectively, for some rank $0 < P \leq D$. The distributions $p(\theta_1)$ and $p(\theta_2)$ are priors on the BNN weights, hence their properties are known or designable a-priori. The function $B(t)$ is Brownian motion implied by a Wiener process $W(t)$ with zero mean and unit covariance without loss of generality, and the related operation around $L_{\theta_2}(h(t), t)$ is the It\^{o} integral \cite{oksendal}. Note that these BNNs may have shared weights, i.e. $\theta_1 \cap \theta_2 \neq \emptyset$. The dynamics of the resultant stochastic process are given by the below stochastic differential equation
\begin{align*}
dx(t) = m_{\theta_1}(x(t),t)dt+L_{\theta_2}(x(t), t)dW(t).
\end{align*}
The process $p(h(t)|\theta_1,\theta_2)$ does not have a closed-form solution, sometimes does not even have an expressable density function, generalizeable to the neural net architecture. However, it is possible to take approximate samples from it by a discretization rule such as Euler-Maruyama. As a work-around, we first marginalize the stochastic process out of the likelihood by Monte Carlo integration
\begin{align*}
p(y|\theta_1,\theta_2,x) = \int p(y| h(T), \theta_1,\theta_2, x) p(h(T)|x) dh(T) \approx \dfrac{1}{M} \sum_{m=1}^M p(y|\tilde{h}_m^T, \theta_1, \theta_2, x),
\end{align*}
where $\tilde{h}_m^T$ is the $T$th time point realization of the Euler-Maruyama draw $m$. This approximation appears in the literature as the simulated likelihood method \cite{applied_sde}. Having integrated out the stochastic process, the rest is a plain approximate posterior inference problem on $p(\theta_1,\theta_2|x,y)$. The sample-driven solution to the stochastic process $h(T)$ integrates naturally into a Markov Chain Monte Carlo (MCMC) scheme. We choose Stochastic Gradient Langevin Dynamics (SGLD) \cite{sgld} with a block decay structure \cite{psgld} to benefit from the loss gradient. Our training scheme is detailed in Algorithm \ref{alg:sgld_block}.

\begin{algorithm}[H] 
	\caption{DBNN Inference}\label{alg:sgld_block}
	\small
	\begin{algorithmic}
		\STATE {\bf Inputs:} Initial weights $\theta^0 := (\theta_1^0, \theta_2^0)$, Decay rate $\lambda$, Flow time $T$, Minibatch size $K$, Iteration count $I$
		\STATE {\bf Outputs:} BNN weights $\{\theta^i\}_{i=1:I}$
		
		\FOR{$i\leftarrow 1 : I$}
		\STATE Sample minibatch  $\{ x_k, y_k \}_{k=1:K}$
		\FOR{$k\leftarrow 1 : K$}
		\STATE $h_{km}^0 = x_k$
		\FOR{$m \leftarrow 0:M$} 
		\FOR{$t \leftarrow 0:T$}
		\STATE $ \tilde{h}_{km}^{t+1} \leftarrow \tilde{h}_{km}^t + m_{\theta_1} ( \tilde{h}_{km}^t, t)\Delta t+L_{\theta_2}( \tilde{h}_{km}^t, t)\Delta W$
		\ENDFOR
		\ENDFOR
		\STATE $\tilde{p}(y_k|\theta_1^{i-1}, \theta_2^{i-1}, x_k) \leftarrow \frac{1}{M}\sum_{m=1}^{M} p(y_k|\tilde{h}_{km}^T,\theta_1^{i-1},\theta_2^{i-1}, x_k)$
		\ENDFOR		
		\STATE $\theta^{i} \leftarrow \theta^{i-1}  + \frac{\epsilon}{2} \left[
		{\nabla \log p(\theta^{i-1}) + \frac{N}{K} \sum_{k=1}^{K} \nabla \log \tilde{p}(y_k|\theta_1^{i-1}, \theta_2^{i-1}, x_k)} \right] 
		+ \mathcal{N}(0, \epsilon)$
		\IF{$ i \mod \lambda = 0$}
		\STATE $\epsilon \leftarrow \epsilon / 2$
		\ENDIF
		\ENDFOR
	\end{algorithmic}
\end{algorithm}

\section{Experiments}
We compare our method DBNN against Neural SDE (N-SDE) \cite{nsde}, its closest and most-recent relative, which applies a fixed-rate dropout on the neural net of the diffusion matrix and uses RMSE as loss function. Hence, we evaluate how making both drift and diffusion neural nets fully Bayesian and using a modified variant of SGLD for posterior weight inference improves the results.

\begin{figure}[b]%
	\centering
	\subfloat[Time series prediction based on noisy and equally spaced observations. Underlying ground truth function is the stochastic Vasicek model.]
	{{\includegraphics[width=6.5cm]{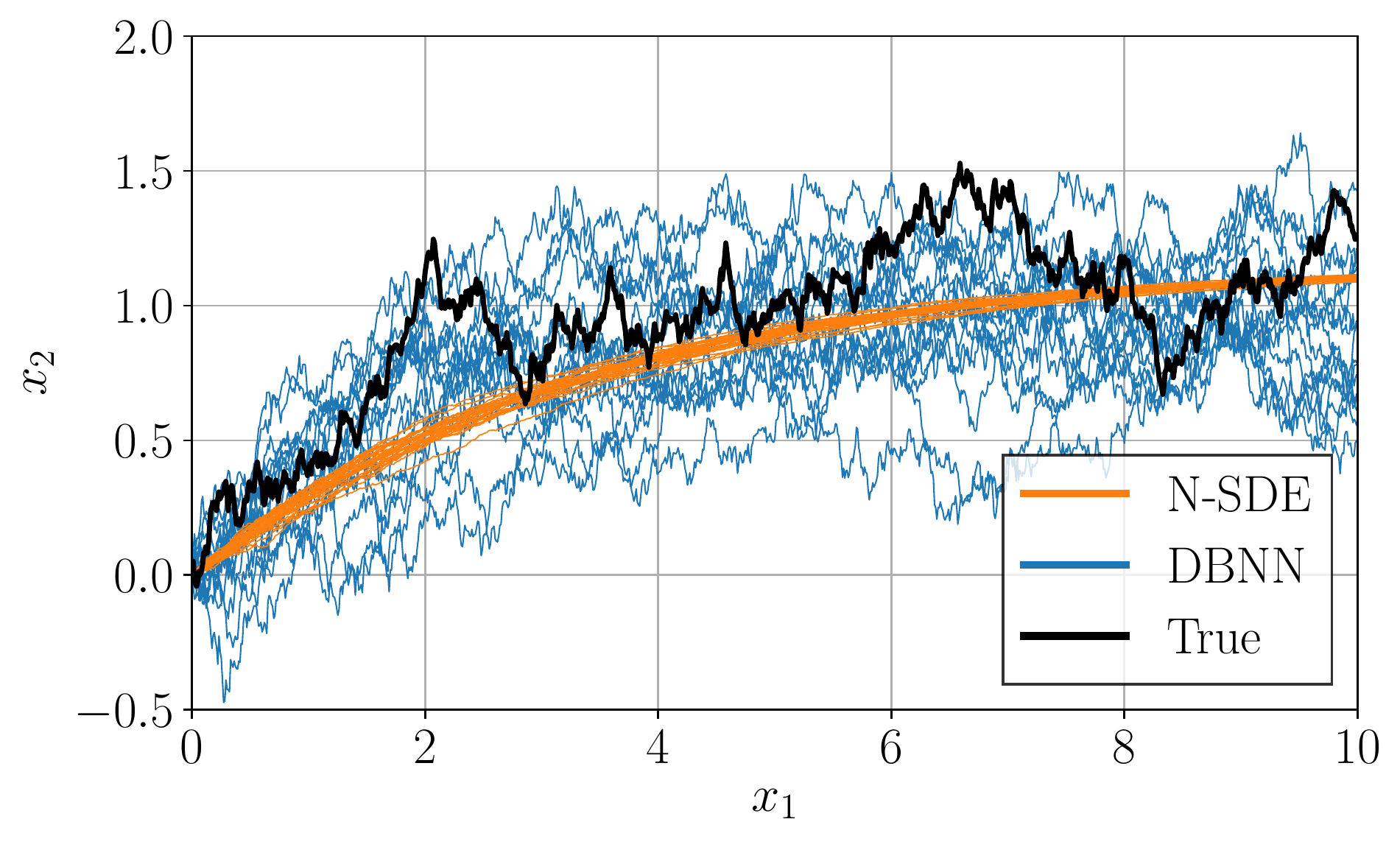} }	\label{fig:time_a}}%
	\qquad
	\subfloat[Time series prediction based on noisy and randomly distributed observations. Underlying ground truth function is the centered sigmoid function.]
	{{\includegraphics[width=6.5cm]{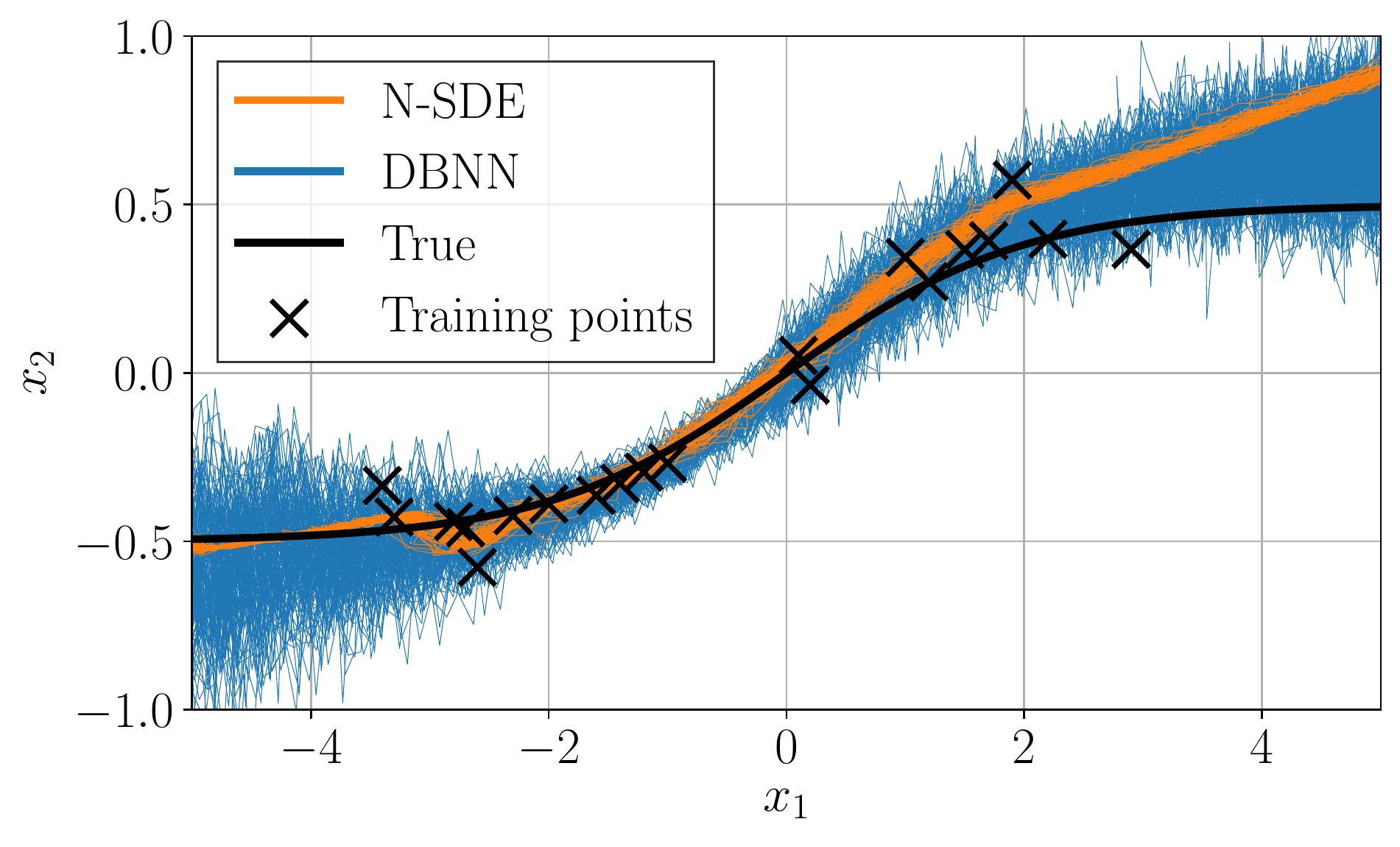} \label{fig:time_b} }}%
	\caption{Time series prediction results for DBNN and N-SDE with fixed dropout diffusion \cite{nsde}.}%
\end{figure}

\paragraph{Time series modeling.} In the first experiment one draw of the Vasicek model \cite{vasicek} with equally spaced observations is given. We specify the model as $dx = 0.5(1-x)dt + 0.25dW$. It starts from the initial point (0,0), converges to 1, and then oscillates around it. Figure \ref{fig:time_a} plots the results. Our method is capable of modeling the underlying dynamics and reflects the noisy nature of the data. In contrast, the NSDE approach results in excessively smooth predictions and uncalibrated uncertainty scores. Figure \ref{fig:time_b} shows results for non-equally spaced data. Ground truth is the centered sigmoid function, from which 20 noisy observations have been sampled. DBNN is capable of representing the predictive uncertainty and shows increasing uncertainty in the interpolation and extrapolation areas. Although N-SDE learns an accurate predictive mean, its uncertainty scores show little to no correlation to the observed data. In both experiments we observed that N-SDE did not converge for dropout rates $>5\%$. Additionally we found N-SDE to behave sensitive towards the choice of time dependence for drift and diffusion. Our results demonstrate the necessarity to properly account for uncertainty during training, as we do in DBNN, in order to get well calibrated predictive uncertainty. 

\paragraph{Regression.} For regression, we place an additional linear layer above $h(T)$ in order to match the output dimensionality. Since we can estimate the properties of the distribution $p(h(T)|x)$, with mean $m_{\theta_1}$ and covariance $L_{\theta_2} L_{\theta_2}^T=\Sigma_{\theta_2}$, we propagate both moments through the linear layer. The predictive mean is thus modeled as $\sum a_i m_{\theta_1, i} + b_i$ and predictive variance as  $\sum a_i a_j \Sigma_{\theta_2, i,j}$. It is possible to design $L_{\theta_2}$ as a diagonal matrix assuming uncorrelated activation map dimensions. Further, $L_{\theta_2}$ can be parameterized by assigning the DBNN output on its Cholesky decomposition, or it can take any other structure of the form $\mathbb{R}^{D \times P}$. When choosing $P < D$, it is possible to heavily reduce the number of learnable parameters. Table \ref{tab:uci_res} shows results for the UCI benchmark dataset. We use the experiment setup (network architecture and train/test splitting schemes) defined in \cite{pbp}. Further, we choose the hyperparameters of N-SDE as in \cite{dropout}. DBNN brings either improved or competitive fit on test data in all data sets. Modeling correlated noise also improves the results in most data sets.  

\begin{table*}[th]
	\caption{Test log likelihood values of 8 benchmark datasets.}
	\label{tab:uci_res}
	
	\resizebox{1.00\textwidth}{!}{
		\begin{tabular}{ l c  cc cc cc cc }
			\toprule
			& & boston & energy & concrete & wine\_red & kin8mn & power & naval & protein \\
			\cmidrule(lr){3-10}
			& $N$ & 506 & 768 & 1,030 & 1,599 & 8,192 & 9,568 & 11,934 & 45,730  \\
			& $D$ & 13 & 8 & 8 & 22 & 8 & 4 & 26 & 9  \\
			\midrule
			
			PBP  \cite{pbp} && -2.57(0.09) & -2.04(0.02) & -3.16(0.02)& -0.97(0.01)& 0.90(0.01)& -2.84(0.01)&
			3.73(0.01)& -2.97(0.00)\\
			Dropout \cite{dropout} && -2.46(0.06) & -1.99(0.02) & \textbf{-3.04(0.02)}& -0.93(0.01) & 0.95(0.01) &
			\textbf{-2.80(0.01)}& 3.80(0.01)& -2.89(0.00)\\
			
			N-SDE \cite{nsde} & Dropout& -2.48(0.03) & -1.35(0.01) &  -3.05(0.03) &-0.97(0.01) & 0.94(0.02) & -2.82(0.01)  & \textbf{3.83(0.03)} & -2.89(0.00) \\
			
			DBNN  & Diagonal & -2.47(0.04) & -1.60(0.09) &  -3.05(0.03) &  -0.93(0.02) & 1.06(0.01) &  -2.81(0.01) & 2.78(0.00) & -2.85(0.01) \\
			
			DBNN & Cholesky &\textbf{-2.45(0.03)} & \textbf{-1.22(0.05)} &  -3.05(0.03) &\textbf{-0.92(0.02)} & \textbf{1.08(0.01)} & \textbf{-2.80(0.00)}  & 2.97(0.09) &  \textbf{-2.81(0.00)} \\
			
			\bottomrule
		\end{tabular}
	}	
\end{table*}

\section{Conclusion}
We extend Neural ODEs to a fully Bayesian setting. The model flows an input observation through stochastic dynamics, where both the drift and the diffusion follow a BNN. The posterior on the BNN weights is approximated by a modified variant of SGLD. The resultant model, called DBNN, outperforms the recent N-SDE in a number of time series prediction and regression tasks.
Our model benefits from the natural flexibility of using a variety of possible network designs for $m_{\theta_1}$ and $L_{\theta_2}$, as long as the input and output dimensions remain same. Thus the model is easily extendable towards other tasks, such as image segmentation or reinforcement learning.

\small
\bibliographystyle{abbrv}
\bibliography{dbnn}

\end{document}